\pdfoutput=1

\documentclass[11pt]{article}

\usepackage{acl}

\usepackage{times}
\usepackage{latexsym}

\usepackage[T1]{fontenc}

\usepackage[utf8]{inputenc}

\usepackage{amsmath}
\usepackage{amsfonts}
\usepackage{amssymb}

\usepackage{subcaption}

\usepackage{booktabs}
\usepackage{microtype}

\usepackage{inconsolata}

\usepackage{graphicx}

\title{DTW-Align: Bridging the Modality Gap in End-to-End Speech Translation with Dynamic Time Warping Alignment}

\author{Abderrahmane Issam \qquad {\bf Yusuf Can Semerci} \qquad {\bf Jan Scholtes} \qquad {\bf Gerasimos Spanakis} \\
        Department of Advanced Computing Sciences \\ 
        Maastricht University \\ 
        \small{\texttt{\{abderrahmane.issam, y.semerci, j.scholtes, jerry.spanakis\}@maastrichtuniversity.nl}}}

\begin{document}
\maketitle
\begin{abstract}

End-to-End Speech Translation (E2E-ST) is the task of translating source speech directly into target text bypassing the intermediate transcription step. The representation discrepancy between the speech and text modalities has motivated research on what is known as \textit{bridging the modality gap}. State-of-the-art methods addressed this by aligning speech and text representations on the word or token level. Unfortunately, this requires an alignment tool that is not available for all languages. Although this issue has been addressed by aligning speech and text embeddings using nearest-neighbor similarity search, it does not lead to accurate alignments. In this work, we adapt Dynamic Time Warping (DTW) for aligning speech and text embeddings during training. Our experiments demonstrate the effectiveness of our method in bridging the modality gap in E2E-ST. Compared to previous work, our method produces more accurate alignments and achieves comparable E2E-ST results while being significantly faster. Furthermore, our method outperforms previous work in low resource settings on 5 out of 6 language directions. \footnote{\scriptsize\texttt{\href{https://github.com/issam9/DTW-Align}{https://github.com/issam9/DTW-Align}}}

\end{abstract}

\section{Introduction}

End-to-End Speech Translation (E2E-ST) is the task of translating speech in a source language directly into text in a target language. E2E-ST gained success and attention as an alternative to cascaded solutions where an Automatic Speech recognition (ASR) and a Machine Translation (MT) models are combined \cite{tang-etal-2021-improving,  ye-etal-2022-cross, fang-etal-2022-stemm, ouyang-etal-2023-waco, zhou-etal-2023-cmot, le2023pretrainingspeechtranslationctc, zhang2024-salign, zhang-etal-2025-representation}. Cascaded solutions benefit from abundant ASR and MT data but might suffer from error propagation and high latency, which can be solved by E2E-ST.

\begin{figure}[ht]
    \centering
    \includegraphics[width=0.49\textwidth]{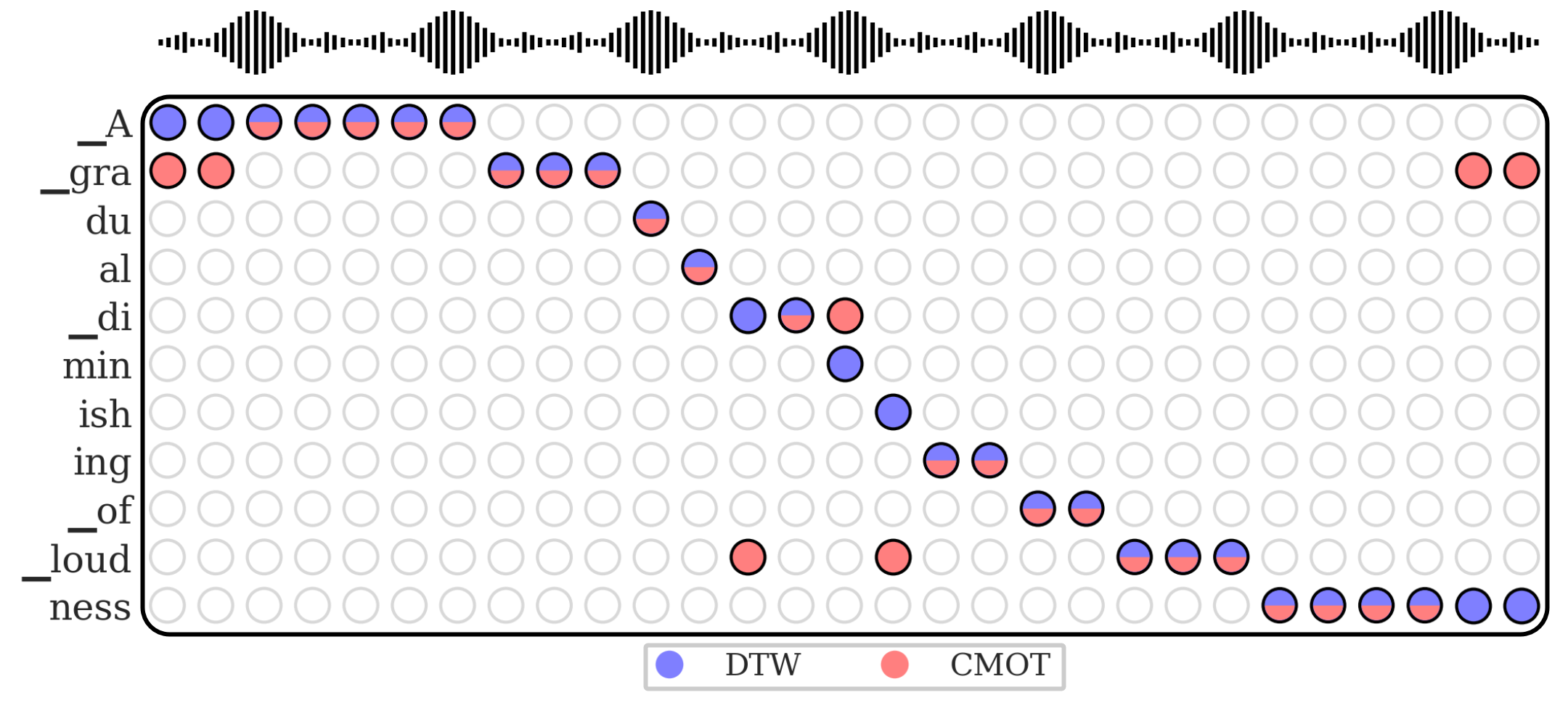}
    \caption{We show an example alignment from DTW (Ours) vs. CMOT. The figure shows that unlike CMOT, DTW guarantees generating monotonic alignments and that all tokens are aligned. In contrast, CMOT failed to align the tokens "min" and "ish" to any frames.}
    \label{fig:dtw_intro}
\end{figure}

However, training E2E-ST models is not straightforward due to the representation discrepancy between the speech and text modalities. Previous work has achieved state-of-the-art results by aligning speech and text representations at the word or token level, either using an alignment tool \cite{ouyang-etal-2023-waco, fang-etal-2022-stemm} or by generating the alignment automatically during training \cite{zhou-etal-2023-cmot, hao2023}. The closest to our work is Cross-modal Mixup via Optimal Transport (CMOT), which uses optimal transport for finding speech and text alignments. Although CMOT achieves state-of-the-art results, it does not guarantee producing monotonic alignments or ensure that each text token is assigned to at least one frame. This contradicts the expected structure of speech-text alignment and can lead to noisy alignments. Furthermore, CMOT introduces a significant training time overhead.

In this work, we introduce \textbf{DTW-Align}, a method for aligning speech and text embeddings during training using an adaptation of Dynamic Time Warping \cite{sakoe1978-dtw}. Figure \ref{fig:dtw_intro} shows an example alignment generated using DTW-Align and CMOT, which illustrates that our method generates monotonic alignments and guarantees that all tokens are aligned, while CMOT does not. We demonstrate the effectiveness of our method in bridging the modality gap with mixup training \cite{fang-etal-2022-stemm, zhou-etal-2023-cmot}. Similarly to previous work \cite{zhou-etal-2023-cmot}, we train on a mixup of aligned speech and text representations, however, instead of discretely selecting either a speech or a text embedding, we linearly interpolate speech and text embeddings \cite{zhang2018mixup}. Our experiments show that our method is faster and produces more accurate alignments. Furthermore, it achieves comparable results to CMOT on 6 language directions from the CoVoST2 dataset, while training significantly faster. We also evaluate our method in a low resource setting where training can be more vulnerable to alignment noise, and we show that our method leads to a statistically significant improvement over CMOT in 5 out of 6 language directions.

\section{Related Works}
\textbf{Bridging the Modality Gap}: 
The discrepancy between the source and target modalities (i.e. speech and text respectively) has motivated multiple works on what is termed bridging the modality gap \cite{yuchen2019,han-etal-2021-learning,fang-etal-2022-stemm}, where the goal is to build a shared semantic space between the speech and text modalities. Aligning speech and text either based on an alignment tool \cite{fang-etal-2022-stemm, ouyang-etal-2023-waco} or dynamically during training \cite{hao2023, zhou-etal-2023-cmot} was shown to achieve state-of-the-art results. Our work goes in this direction, by improving the accuracy and speed of aligning speech and text during training.

\noindent
\textbf{Mixup:}
Mixup is a common data augmentation strategy \cite{zhang2018mixup, jin2025surveymixupaugmentations}. In E2E-ST, it is applied for bridging the modality gap \cite{fang-etal-2022-stemm, zhou-etal-2023-cmot}, where the model is trained on a discrete mixup of speech and text representations. Mixup training in E2E-ST requires an alignment between speech and text that can be generated using an alignment tool \cite{fang-etal-2022-stemm}. \citet{zhou-etal-2023-cmot} alleviate the need for an alignment tool by aligning speech and text representations using optimal transport. Our approach is similar to \cite{zhou-etal-2023-cmot}, where we generate the alignments dynamically during training. However, instead of discretely mixing speech and text representations, we apply mixup as a linear interpolation.

\noindent
\textbf{DTW:}
DTW is an algorithm for measuring similarity between two sequences of varying length \cite{sakoe1978-dtw}. Due to this property, it has been widely applied to speech data \cite{juang1984-dtw, furtuna2008, muda2010-dtw}, and also more specifically in the context of aligning speech and text sequences (i.e. forced alignment). For example, Aeneas \cite{pettarin2017aeneas} aligns speech and text utterances by transforming the text utterances into speech, then uses DTW to align the synthetic and original speech sequences. \citet{ludwig2020-ctc} uses an algorithm that resembles DTW by using dynamic programming and backtracking to find the optimal alignment based on Connectionist Temporal Classification (CTC) probabilities. In this work, we adapt DTW to dynamically align speech and text based on their embeddings.

\section{Method}
\subsection{Architecture}

Inspired by previous work in E2E-ST \cite{fang-etal-2022-stemm, zhou-etal-2023-cmot}, our model consists of two main components, a speech encoder, and a translation encoder-decoder. The translation encoder-decoder is a standard transformer model that can be decomposed into 3 components: a text embedding layer, an encoder that inputs either speech or text embeddings, and a decoder that generates the target sentence.

\subsection{DTW for Aligning Speech and Text Representations}
\label{sec:dtw}
DTW can be used to compute similarities between two sequences of variable length along time. This is achieved by finding an optimal path between the two sequences, or the path that leads to their maximum similarity. The time dimension of the two sequences is said to be \textit{warped}. In our case, when aligning speech and token embeddings, we only warp the token time dimension to have a one-to-many relationship from speech to token embeddings. We start by computing the cosine similarity between each speech embedding $t\in [0;N-1]$ to each token $j \in [0;M-1]$, then we use the similarity matrix $S \in \mathbb{R}^{N \times M}$ to compute a trellis matrix $T$ of the same dimension:

{\small
\begin{equation}
\begin{aligned}
T_{t,j} = 
\begin{cases}
S_{t, j} &  t=0, j=0 \\
-\infty &  t=0, j>0 \\
+\infty & t>N-M, j=0 \\
S_{t, j} + T_{t-1, j} &  t>0, j=0 \\
\max(T_{t-1, j}, T_{t-1, j-1}) \\ + S_{t, j}  & t>0, j>0
\end{cases}
\end{aligned}
\end{equation}
}

The last step is backtracking, where we traverse the trellis starting from the last frame and token to find the optimal path, or the path with maximum similarity, which eventually represents the alignment $a$ from speech to text tokens. We assign the last token to the last frame $a_{N-1}=M-1$, and we traverse as follows:

{\small
\begin{equation}
\begin{aligned}
a_{t} = 
\begin{cases}
    M-1 & t=N-1 \\
    a_{t+1} - 1 & T_{t, a_{t+1}-1} > T_{t, a_{t+1}} \\          
    a_{t+1} & \text{else}                                       
\end{cases}
\end{aligned}
\end{equation}
}

\begin{figure}[t!]
    \centering
    \includegraphics[width=0.25\textwidth]{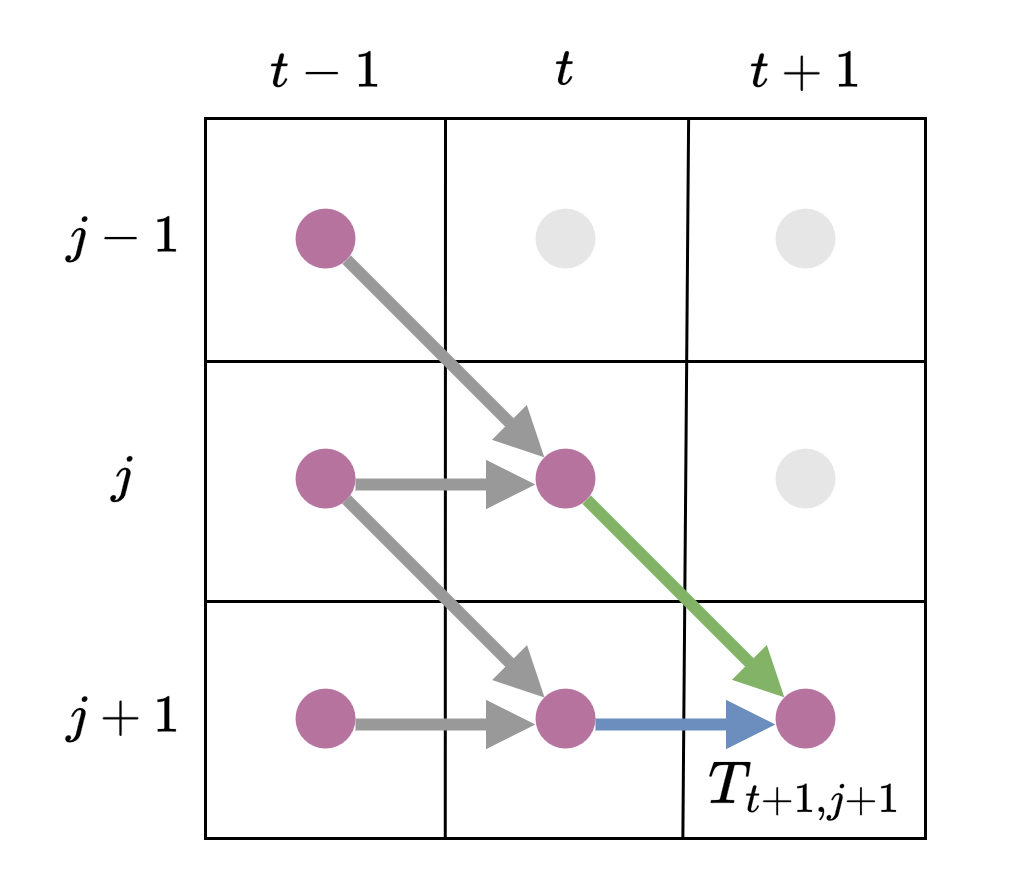}
    \caption{An illustration of the possible alignment paths. Each frame is assigned to only one token, while a token can be assigned to multiple frames.}
    \label{fig:dtw_method}
\end{figure}
\noindent
Figure \ref{fig:dtw_method} shows an illustration of the possible alignment paths. We can see that when backtracking from $t+1,j+1$ we can either go to the previous token $j$ if $T_{t,j}>T_{t,j+1}$ or stay on token $j+1$ otherwise, which guarantees monotonicity. The constraints in the trellis matrix guarantee that all tokens are aligned to at least one frame, since the diagonal is filled with $-\infty$ during the trellis computation, when backtracking, this guarantees that we move to $j-1$ when $j\ge t$. 

By fully vectorizing both the trellis computation and backtracking, our implementation achieves a much faster alignment.

\subsection{Mixup Training}
\label{sec:mixup}
Given a sequence of speech representations generated using the speech encoder $f=[f_0, f_1, ..., f_{N-1}]$ and a sequence of text embeddings generated using the text embedding layer $e=[e_0, e_1, ..., e_{M-1}]$, our method generates an alignment $a=[a_0,a_1,...,a_{N-1}]$ as described in  \S\ref{sec:dtw}. Finally, we apply mixup similarly to previous work \cite{zhou-etal-2023-cmot}:

{\small
\begin{equation}
\begin{aligned}
m_i = 
\begin{cases}
    f_i & p > p^* \\
    e_{a_i} & else
\end{cases}
\end{aligned}
\label{eq:discrete}
\end{equation}
}
where $p^*$ is the mixup probability which controls how much text embeddings we introduce into the speech manifold, and $p$ is sampled from a uniform distribution $\mathcal{U}(0, 1)$. We term this discrete mixup. We further introduce interpolation mixup \cite{zhang2018mixup}, where instead of selecting a speech or text embedding based on probability $p^*$, we use $p^*$ as a mixup coefficient to linearly interpolate speech and text embeddings:

{\small
\begin{equation}
m_i = (1-p^*).f_i + p^*.e_{a_i}
\label{eq:interpolation}
\end{equation}
}
We argue that interpolation mixup can be more robust to alignment noise since the speech embeddings are not entirely replaced as in discrete mixup, but they are softly down-weighted. Therefore, even in the presence of alignment noise, the model still has access to the correct speech embeddings. Furthermore, it can be more data efficient, since all the speech and text token embeddings are included in training, rather than selecting one or the other. 

\subsection{Training Objective}
We train with similar training objectives as CMOT \cite{zhou-etal-2023-cmot} to ensure fair comparison. The ST training corpus is denoted as $D={(s,x,y)}$, where $s$ is the speech input, $x$ is the transcription, and $y$ is the translation. In the first stage, the translation encoder-decoder is pre-trained on transcription-translation pairs using cross entropy:

{\small
\begin{equation}
\mathcal{L}_{MT} = -\mathbb{E}_{x,y} \log P(y|x)
\end{equation}
}
The second stage is multi-task fine-tuning with ST and MT tasks using cross entropy:

{\small
\begin{equation}
\begin{aligned}
\mathcal{L}_{ST} = -\mathbb{E}_{s,x,y} \log P(y|s) \\
\mathcal{L}_{MT} = -\mathbb{E}_{s,x,y} \log P(y|x)
\end{aligned}
\end{equation}
}
\noindent
Furthermore, to bridge the modality gap between speech and text representations, we train with Kullback-Leibler (KL) divergence between the output probability distribution under mixup input $m$, and the output distribution of the ST task, as well as with the output distribution of the MT task:

{\small
\begin{equation}
    \begin{aligned}
        \mathcal{L}_{KL_{m \leftrightarrow s}} = \mathbb{D}_{KL}(P(y|s)||P(y|m)) + \\ 
        \mathbb{D}_{KL}(P(y|m)||P(y|s)) 
    \end{aligned}
\end{equation}
}

{\small
\begin{equation}
\begin{aligned}
        \mathcal{L}_{KL_{m \leftrightarrow x}} = \mathbb{D}_{KL}(P(y|x)||P(y|m)) + \\
        \mathbb{D}_{KL}(P(y|m)||P(y|x))
    \end{aligned}
\end{equation}
}
Therefore, the final loss is:
{\small
\begin{equation}
\begin{aligned}
    \mathcal{L} = \mathcal{L}_{ST} + \mathcal{L}_{MT} +
    \lambda.(\mathcal{L}_{KL_{s \leftrightarrow m}}+ \mathcal{L}_{KL_{x \leftrightarrow m}})/2
\end{aligned}
\end{equation}
}
where $\lambda$ is a hyperparameter weight to control the KL losses.
\section{Experiments}
\subsection{Dataset}
We conduct our experiments on CoVoST-2 dataset \cite{wang2020covost}, a large multilingual ST dataset that is based on Common Voice project \cite{ardila-etal-2020-common}. CoVoST-2 covers translation from 21 source languages to English and from English to 15 target languages, and it contains speech, transcription and translation triplets. In this work, due to computational resources, we focus on 6 language directions: En-De, En-Ca, En-Ar, De-En, Fr-En, and Es-En. These directions are selected to ensure a balanced number of En-X and X-En directions. Furthermore, all languages selected are high resource with a minimum of 97 hours of training data and are of varying linguistic distance from English. 

\subsection{Experimental Setup}

\noindent
\textbf{Pre-processing:} \\
For speech input, we use the raw 16 bit 16kHz mono-channel audio. We filter out examples with a number of frames higher than 480k or less than 1k. For the text input, we remove punctuation, then we tokenize using a uni-gram SentencePiece model \cite{kudo-richardson-2018-sentencepiece} with a vocabulary of 10k that is shared between the source and target languages.

\noindent
\textbf{Model:} \\
Our model is composed of a speech encoder and a translation encoder-decoder. For the speech encoder, we use a pre-trained base HuBERT model \cite{hsu-2021-hubert} for En-X language directions, and mHuBERT-147 \cite{zanon-boito2024mhubert} (a multilingual version of HuBERT base model) for X-En language directions. To shrink the audio representations over the time axis, we stack 2 1-dimensional convolution layers of kernel size 5, stride size 2, padding 2, and hidden dimension 1024. For the translation encoder, we use 6 transformer encoder layers. For the translation decoder, we use 6 transformer decoder layers. Each transformer layer is comprised of 512 hidden units, 8 attention heads, and 2048 feed-forward hidden units.

\noindent
\textbf{Training:} \\
We train our model in two stages, first we pre-train the translation encoder-decoder on CoVoST2 transcription-translation pairs. We train with a learning rate of 1e-4, a maximum of 33k tokens per batch, and for a maximum 100k steps. We early stop training if the loss doesn't decrease for 10 epochs. During the second stage, we fine-tune the speech encoder and translation encoder-decoder with a learning rate of 1e-4, a maximum of 16M audio frames per batch, and we train for 40k steps. For CMOT, NFA-Align and DTW-Align, we train with a mixup probability $p^*=0.2$ and a KL weight $\lambda=2.0$.

The MT models are trained using one A100 GPU and ST models are trained using one H100 GPU. We use Fairseq \footnote{\scriptsize \texttt{\href{https://github.com/facebookresearch/fairseq}{https://github.com/facebookresearch/fairseq}}} \cite{ott-etal-2019-fairseq} for the implementation.

\noindent
\textbf{Evaluation:} \\
We average the last 10 epoch checkpoints for evaluation, and generate with a beam size of 5. We use SacreBLEU \cite{post-2018-call} to compute detokenized case-sensitive BLEU score \cite{papineni2002}.  We also use SacreBLEU to measure statistical significance using paired approximate randomization \cite{riezler-maxwell-2005-pitfalls}.

\noindent
\textbf{Low Resource Setting:} \\
All the languages in our experiments are considered high resource with at least 97 hours of training data, therefore, to evaluate our method in a low resource setting, we simulate a low resource scenario by sampling 10 hours of ST training data and 1 hour of development data for each language directions. During training, we use the same hyperparameters but we early stop if the loss did not decrease on the development set for 10 epochs. Our goal is to demonstrate how noise in the alignment has a more pronounced effect in low-resource ST scenarios. Therefore, we use a simulated low-resource setting with the same languages and training setup to avoid any confounding effects that would arise from using a different dataset.

\subsection{Main Results}
\begin{table*}[ht]
    \centering
    \label{tab:model-comparison}
    \begin{tabular}{lccccccc}
    \toprule
        \textbf{Model} & \textbf{En-De} & \textbf{En-Ca} & \textbf{En-Ar} & \textbf{De-En}  & \textbf{Fr-En} & \textbf{Es-En} & \textbf{Avg.} \\
    \midrule
        Revist-ST (\cite{pmlr-v162-zhang22i})\dag & 17.5 & 22.9 & 12.3 & 14.1 & 26.9 & 15.7 & - \\
        U2TT (Large) \cite{zhang-etal-2023-dub}\dag & - & - & - & 16.7 & 27.4 & 28.1 & - \\ 
        DUB (Large) \cite{zhang-etal-2023-dub}\dag & - & - & - & 19.5 & 29.5 & \textbf{30.9} & - \\
        SRPSE \cite{zhang-etal-2025-representation}\dag & - & - & - & 21.4 & 29.3 & - & - \\
        CoVoST-2 \cite{wang2020covost}\dag & 18.4 & 23.6 & 13.9 & 18.9 & 27.0 & 28.0 & 21.6 \\
        CTC+OT \cite{le2023pretrainingspeechtranslationctc}\dag & 20.6 & 26.5 & 15.3 & 20.4 & 28.4 & 29.2 & 23.4 \\
        HuBERT-Transformer & 21.4 & 27.4 & 15.7 & 21.8 & 28.4 & 28.0 & 23.8 \\ 
        CMOT & \textbf{21.8} & 28.2 & \textbf{16.2} & 23.6 & 30.9 & 29.6 & \textbf{25.0} \\ 
        NFA-Align & 21.4 & 28.1 & 16.0 & 23.5 & 30.9 & 29.8 & 24.9 \\ 
        DTW-Align-Discrete & 21.7 & \textbf{28.3} & \textbf{16.2} & 23.5 & \textbf{31.0} & 29.4 & \textbf{25.0} \\ 
        DTW-Align & \textbf{21.8} & 28.2 & 16.1 & \textbf{23.7} & 30.8 & 29.5 & \textbf{25.0} \\ 
    \bottomrule
    \end{tabular}
    \caption{BLEU score results on CoVoST2 test set. The table shows that CMOT, DTW-Align, and DTW-Align-Discrete achieve the best results against other baselines. \dag: indicates results reported in the original work (the rest of the baselines are trained in this study)}
    \label{table:main_results}
\end{table*}

\noindent
\textbf{Baselines:} \\
\noindent
We experiment with the following models:\\
\noindent
\textit{\textbf{HuBERT-Transformer:} }
Composed of speech encoder and translation encoder-decoder trained for ST. \\
\textit{\textbf{CMOT:}}
HuBERT-Transformer trained by using CMOT alignment for discrete mixup training. \\
\textit{\textbf{NFA-Align:}}
Using word level alignments from NeMo Forced Aligner (NFA) \footnote{\scriptsize \texttt{\sloppy\url{https://github.com/NVIDIA/NeMo/tree/main/tools/nemo\_forced\_aligner}}} which was shown to achieve state-of-the-art results in terms of alignment accuracy \cite{rastorgueva2023nemo} for mixup training. \\
\textit{\textbf{DTW-Align-Discrete (Ours):}}
Using DTW for generating alignments and training with discrete mixup (Equation \ref{eq:discrete}) similar to CMOT. \\
\textit{\textbf{DTW-Align (Ours):}}
Using DTW for generating alignments and training with interpolation mixup (Equation \ref{eq:interpolation}). \\

\section{Results and Discussion}

Table \ref{table:main_results} shows the results of our method and baselines, and results of previous work that was evaluated on the CoVoST2 dataset. The results show that consistent with previous studies \cite{fang-etal-2022-stemm}, the baseline HuBERT-Transformer remains a competitive baseline, even outperforming previous work that uses more complex techniques. Furthermore, CMOT, DTW-Align-Discrete and DTW-Align achieve the best results overall. Although we train under similar settings and we do not optimize our method differently, we achieve similar results to CMOT. Surprisingly, NFA-Align which uses NFA to align speech and text lags slightly behind on average (i.e. 0.1 BLEU), this suggests that in a high resource setting, and with a low mixup probability the effect of noise in the alignment is less evident.

\subsection{Alignment Accuracy and Training time}
\begin{table}[ht]
\scriptsize
    \centering
    \begin{tabular}{llcccc}
        \toprule
        & \textbf{Method} & \textbf{Accuracy} $\uparrow$ & \textbf{Execution Time} $\downarrow$ & \textbf{Train Time} $\downarrow$ \\
        \midrule
        & CMOT & 26\% & 97.89 & 14:20:53 \\
        & DTW-Align & \textbf{45\%} & \textbf{2.91} & \textbf{6:48:14} \\
        \bottomrule
    \end{tabular}
    \caption{We show the accuracy of alignments against NFA, and the execution time on CoVoST2 En-De dev set, plus the training time on En-De train set. DTW-Align is significantly faster and more accurate than CMOT.}
    \label{table:acc_time}
\end{table}

Table \ref{table:acc_time} shows that our method produces more accurate alignments with a significant increase of 19\% in alignment accuracy. Furthermore, our method is more than 33 times faster in terms of execution time, which is concretely manifested in the staggering difference in training time between CMOT and DTW-Align (i.e. 14:20:53 and 6:48:14 respectively). As a reference, HuBERT-Transformer baseline training time is 6:32:53, which means that our method improves the performance over this baseline (by an average of 1.2 BLEU points) without the drawbacks of the significant training time overhead that CMOT suffers from. Therefore, although our method achieves similar results to state-of-the-art CMOT in high resource settings, it offers a significant advantage in terms of training time. In Section \ref{analysis}, we show that due to the improved alignment accuracy, our method is more robust both in low resource settings and under higher mixup probability values.

\section{Analysis}
\begin{table*}[ht]
    \centering
    \label{tab:model-comparison}
    \begin{tabular}{lccccccc}
    \toprule
        \textbf{Model} & \textbf{En-De} & \textbf{En-Ca} & \textbf{En-Ar} & \textbf{De-En}  & \textbf{Fr-En} & \textbf{Es-En} & \textbf{Avg.} \\
    \midrule
        HuBERT-Transformer & 6.4 & 8.7 & 2.2 & 1.8 & 3.2 & 2.9 & 4.2 \\ 
        CMOT & 6.6 & 9.6 & 2.7 & 2.8 & 8.5 & 7.5 & 6.3 \\ 
        DTW-Align-Discrete & 6.8** & 9.6 & 2.8 & 2.7 & \textbf{8.6} & 7.9** & 6.4 \\ 
        DTW-Align & \textbf{7.0**} & \textbf{9.8**} & \textbf{3.1**} & \textbf{2.9*} & \textbf{8.6} & \textbf{8.0**} & \textbf{6.6} \\ 
    \bottomrule
    \end{tabular}
    \caption{BLEU score results on CoVoST2 test set in the low resource setting. The table shows that on overall DTW-Align-Discrete and DTW-Align on overall achieve better results than CMOT, with DTW-Align achieving the best results overall. *, ** indicate whether the improvement over CMOT is statistically significant with $p<0.05$ and $p<0.01$ respectively.}
    \label{table:lrs_results}
\end{table*}
\begin{figure*}
    \centering
    \begin{subfigure}{0.45\textwidth}
        \includegraphics[width=1.0\textwidth]{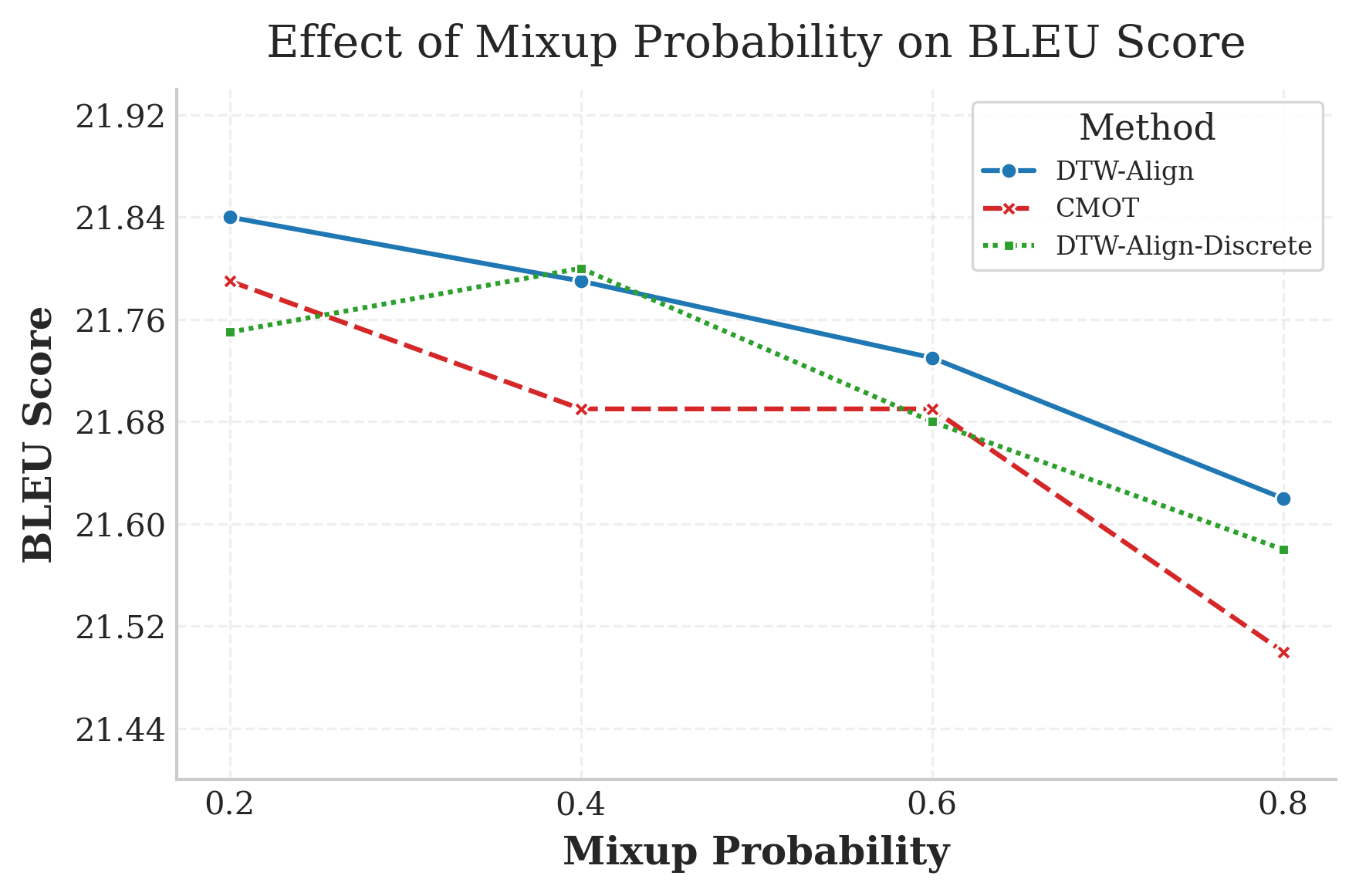}
    \caption{En-De}
    \label{fig:high}
    \end{subfigure}
    \begin{subfigure}{0.45\textwidth}
        \includegraphics[width=1.0\textwidth]{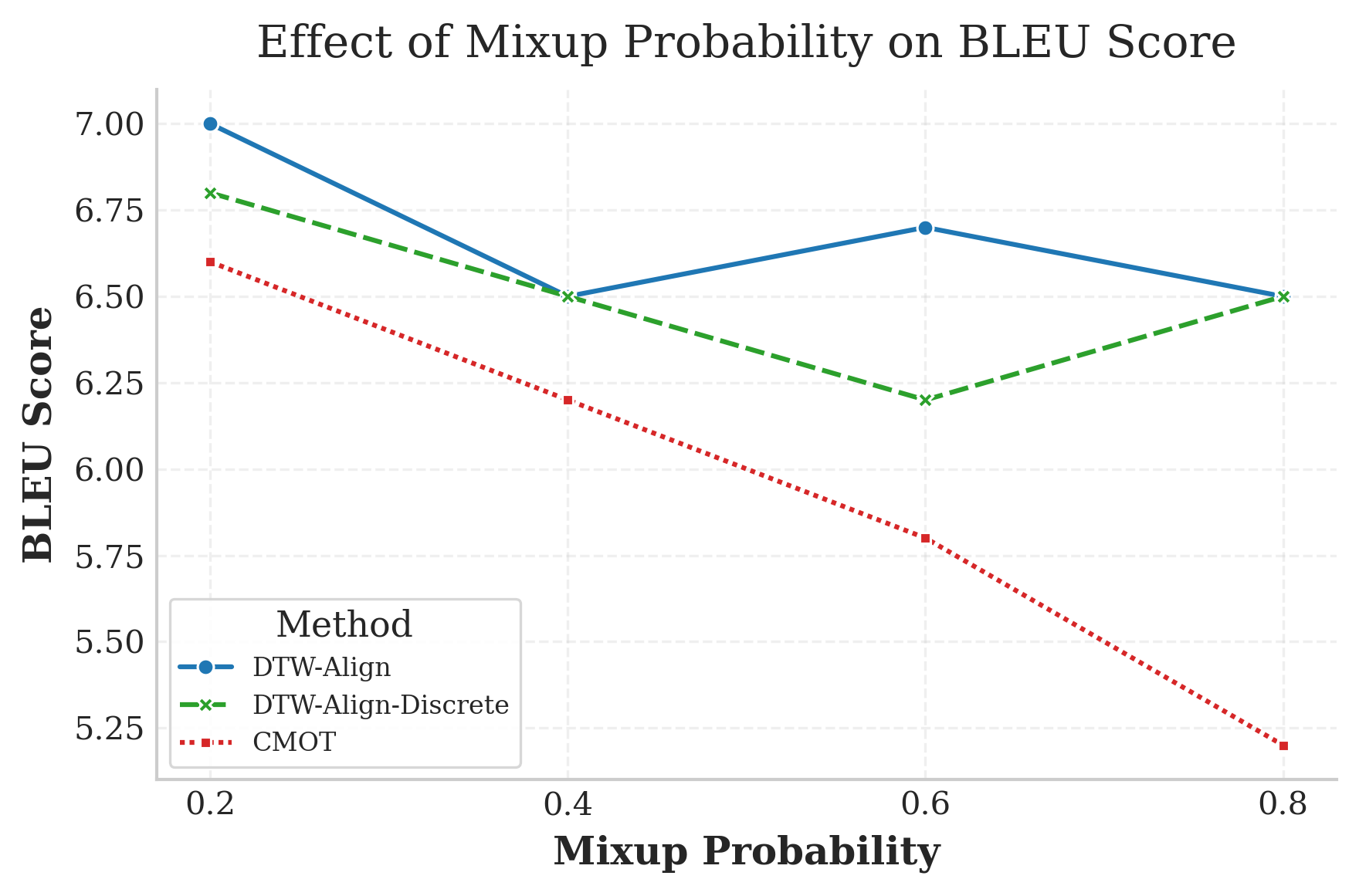}
    \caption{En-De Low Resource}
    \label{fig:low}
    \end{subfigure}
    \caption{The BLEU score of CMOT and DTW-Align under different mixup probabilities on En-De (Figure \ref{fig:high}) and En-De Low Resource (Figure \ref{fig:low}). DTW-Align is more robust to higher mixup probabilities than CMOT even with discrete mixup. This can be explained by noise in CMOT alignments.}
    \label{fig:mixup_prob}
\end{figure*}
\label{analysis}
Although our method substantially outperfroms CMOT in terms of alignment accuracy, it does not yield improvements in ST performance. We attribute this to two factors: the amount of training data, which makes training more robust under noise and the low mixup probability value, which is set to 0.2. In \S\ref{sec:low_res} we measure the performance of CMOT, DTW-Align-Discrete and DTW-Align in a simulated low resource scenario of 10h per language direction, and in \S\ref{sec:mixup_prob} we ablate the mixup probability value.
\subsection{Low Resource Setting}
\label{sec:low_res}

Models can be more vulnerable to the negative effects of alignment noise in low resource scenarios. To study this, we compare the performance of CMOT, DTW-Align-Discrete and DTW-Align in a low resource setting of 10h of ST training data and 1h of development data. Table \ref{table:lrs_results} shows the results over the 6 language directions in our experiments. Overall, DTW-Align-Discrete achieves better results than CMOT, with the improvements on En-De and Es-En being statistically significant. Furthermore, DTW-Align achieves the best results, with statistically significant improvement over CMOT on 5 language directions out of 6. These results show that combining the alignment accuracy of DTW and the robustness of interpolation mixup yields the best performance in low resource settings. Although our method performs on par with CMOT in high-resource settings, it offers an increase in performance in low resource ones, where effects of noise on CMOT are more pronounced. Finally, we find that the improvement of DTW-Align over HuBERT-Transformer has doubled (i.e. from 1.2 to 2.4 BLEU points), which demonstrates the advantage of mixup training in low resource settings.

\subsection{Mixup Probability}
\label{sec:mixup_prob}
We perform an ablation study on the effect of increasing the mixup probability $p^*$ of CMOT, DTW-Align-Discrete and DTW-Align as shown in Figure \ref{fig:mixup_prob} on En-De in high (Figure \ref{fig:high}) and low resource setting (Figure \ref{fig:low}). Results indicate that higher mixup probabilities lead to lower performance but the performance degradation is more significant in the case of CMOT, especially in the low resource setting, where training is more vulnerable to noise. This demonstrates that using DTW for aligning speech and text representations is more robust to the mixup probability hyperparameter, especially in low resource scenarios.

\section{Conclusion}
We introduce a method that eliminates the requirement for an external forced alignment tool by dynamically aligning speech and text embeddings during training based on Dynamic Time Warping (DTW). Compared to state-of-the-art approaches, our method matches or exceeds BLEU score results while being significantly faster. We further demonstrate that using DTW-Align is more robust and data efficient in low resource settings. In addition, compared to HuBERT-Transformer baseline, our method improves performance by 1.2 and 2.4 BLEU points in high and low resource settings respectively with minimal overhead in the training time. Finally, unlike CMOT, our method can produce both token and word level alignments, which makes it compatible with previous work that requires word level alignments \cite{fang-etal-2022-stemm, ouyang-etal-2023-waco, nguyen-etal-2025-spirit}, therefore, it can bring a boost to the ongoing efforts on bridging the modality gap in E2E-ST or other speech-to-text tasks.

\section*{Limitations}
Our work considers the following limitations:

Previous work shows that using external MT data for pretraining the translation encoder-decoder improves downstream ST performance. In our experiments, however, we only use internal CoVoST2 data for pretraining because of resource limitations.

Moreover, our work requires speech transcriptions, which might not be available for all languages. Future work can explore using transcriptions from an ASR model potentially extending the method's applicability to a wider range of languages.

Finally, CoVoST2 is an English centric dataset with English as the source or target language in all directions. Evaluating the accuracy and effect of speech and text alignment on other language directions would be valuable for future research.

\section*{Acknowledgments}
The research presented in this paper was conducted as part of VOXReality project\footnote{\texttt{\url{https://voxreality.eu/}}}, which was funded by the European Union Horizon Europe program under grant agreement No 101070521.

This work used the Dutch national e-infrastructure with the support of the SURF Cooperative using grant no. EINF-11297.

\bibliography{custom}

\appendix

\end{document}